\documentclass[10pt,twocolumn,letterpaper]{article}
\usepackage[camera-ready]{cvpr}      

\usepackage{graphicx}
\usepackage{amsmath}
\usepackage{amssymb}
\usepackage{booktabs}
\usepackage{svg}
\usepackage{multirow} 

\usepackage{amssymb} 

\newcommand{\cmark}{\ding{51}}
\newcommand{\xmark}{\ding{55}}

\usepackage{pifont}

\usepackage[pagebackref,breaklinks,colorlinks]{hyperref}

\usepackage[capitalize]{cleveref}
\crefname{section}{Sec.}{Secs.}
\Crefname{section}{Section}{Sections}
\Crefname{table}{Table}{Tables}
\crefname{table}{Tab.}{Tabs.}

\def\cvprPaperID{*****} 
\def\confName{CVPR}
\def\confYear{2022}

\begin{document}

\title{FIRST: A Million-Entry Dataset for Text-Driven  Fashion Synthesis and Design}
\author{Zhen Huang$^{1,2}$, Yihao Li$^1$, Dong Pei$^1$, Jiapeng Zhou$^{1,2}$, Xuliang Ning$^{1,2}$,\\
Jianlin Han$^3$, Yanlin Li$^1$, Xiaoguang Han$^{1,2 *}$, Xuejun Chen$^{1,3 *}$
\and
$^1$ChimerAI Inc\\
$^2$Chinese University of Hong Kong, Shenzhen\\
$^3$Huizhou University
{
}
}


\maketitle

\begin{abstract}
Text-driven fashion synthesis and design is an extremely valuable part of artificial intelligence generative content(\textbf{AIGC}), which has the potential to propel a tremendous revolution in the traditional fashion industry. To advance the research on text-driven fashion synthesis and design, we introduce a new dataset comprising a million high-resolution \textbf{f}ashion \textbf{i}mages with  \textbf{r}ich \textbf{s}tructured \textbf{t}extual(\textbf{FIRST}) descriptions. In the FIRST, there is a wide range of attire categories and each image-paired textual description is organized at multiple hierarchical levels. Experiments on prevalent generative models trained over FISRT show the necessity of FIRST. We invite the community to further develop more intelligent fashion synthesis and design systems that make fashion design more creative and imaginative based on our dataset. The dataset will be released soon.
\end{abstract}

\begin{figure}
    \centering
    \includegraphics[width=0.90\linewidth]{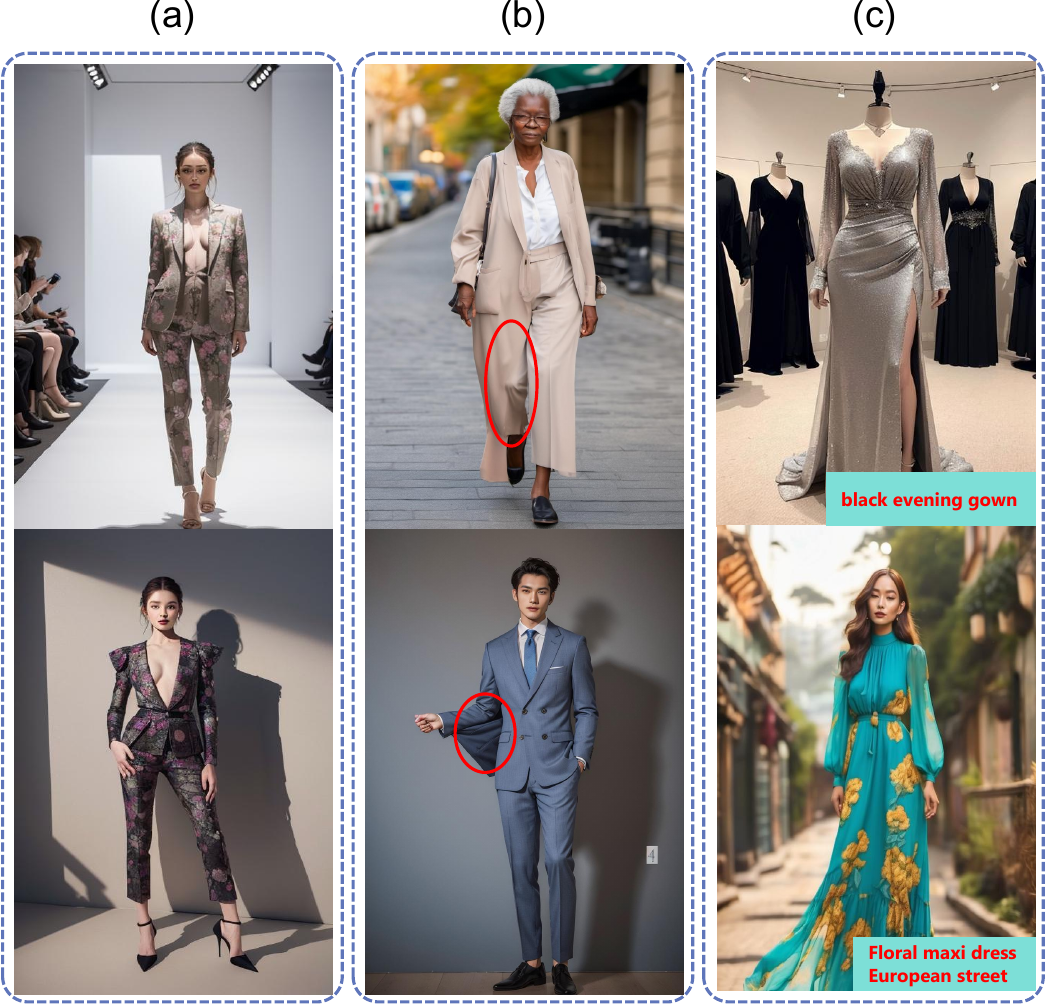}
    \caption{Three major types of failed cases occur in the images produced by SDXL. Column (a) shows the unnatural structure of faces. Column (b) includes the incorrect structure of generated clothes. Column (c) demonstrates the weak controllability of SDXL.}
    \label{problem}
\end{figure}

\section{Introduction}
\let\thefootnote\relax\footnotetext{$^{*}$ Corresponding author}
\label{sec:intro}
Fashion synthesis and design leverage cutting-edge deep learning techniques to revolutionize how we create and visualize garments, offering personalized and scalable solutions to the dynamic demands of the fashion industry\cite{mohammadi2021smart}. In addition to high-generation quality, a great intelligent fashion design system should be convenient to interact with and significantly contribute to emancipating productivity. Naturally, textual description is adopted as an interactive interface for such a system\cite{chen2023survey}. With the advent and evolution of diffusion models, the text-to-image task is paid more attention, and a variety of text-driven generative models\cite{podell2023sdxl,ramesh2022hierarchical,saharia2022photorealistic,ramesh2021zero} have emerged with dramatically increasing power. Among these models, stable diffusion stands out and becomes a prevalent base model for secondary development because of the availability of its code and parameters trained on LAION-5B\cite{schuhmann2022laion}. Nevertheless, the stable diffusion is trained for generic creativity. It does not understand well the complex fashion design elements and abstract concepts, which causes degraded generated results and diminishes text-driven controllability of stable diffusion over images. Figure \ref{problem} shows some failed cases generated by SDXL\cite{podell2023sdxl}. Although the stable diffusion can be efficiently trained on small-scale fashion text-image pairs by using finetune techniques\cite{ruiz2023dreambooth, gal2022image,hu2021lora,ha2016hypernetworks} to reinforce its understanding of the correspondence between fashion concepts and visual elements, such small-scale fashion images with insufficient and unstructured textual descriptions still restrict the model with limited creativity and imagination.

Existing works\cite{liu2016deepfashion,choi2021viton,fu2022stylegan,morelli2022dress} have released some fashion datasets. However, they do not provide textual descriptions for fashion images. While other works\cite{jiang2022text2human,baldrati2023multimodal} annotate images with paired text, they either have simple short texts without the characteristics of human models and backgrounds which both play prominent parts in fashion design or only have small-scale image-text pairs. All aforementioned datasets are inappropriate for fashion design systems.  

To unveil the power of fashion design and encourage the community to construct formidable fashion design systems, we present a large-scale fashion dataset consisting of high-resolution \textbf{f}ashion \textbf{i}mages with \textbf{r}ich \textbf{s}tructured \textbf{t}extual(\textbf{FIRST}) descriptions. The FIRST includes numerous fashion items and styles from world-class designers, which provides large room for generative models to create infinity. To obtain the paired texts,  we first utilize the GPT-4v\cite{openai2023gpt} to hierarchically generate elaborate descriptions according to our carefully designed prompts. Then, we revise generated texts by humans to ensure that there are no subtle descriptive and logical mistakes. We also developed a web application for efficiently distributing the images to annotators. To our best knowledge, this is the first fashion dataset that includes a million image-text pairs.

In this paper, we also propose two challenges based on The FIRST. The first challenge is how to expand diffusion models to take longer text as input. In the FIRST, each textual description is composed of almost a thousand tokens for details. However, The dominant diffusion-based algorithms have a strict limitation on the number of tokens because they are equipped with CLIP\cite{radford2021learning} which requires that the number of input tokens must be 77. Even though SDXL adds another OpenCLIP model as the second text encoder, it still does not gratify the practical demand in the fashion industry. The second challenge is how to synthesize a collection of fashions based on a few references that share the same design philosophy. A collection is a professional term in fashion design. Normally, different fashion categories are present in a collection and they share a portion of fashion elements and concepts. The ability to produce a collection by fashion design systems can significantly reduce laborious design work. We release the FIRST soon and invite more researchers to put their interest in these two challenges.

In summary, our contributions in this paper are as follows:
\begin{itemize}
\item We have introduced the first large-scale fashion generation dataset with a million instances, dubbed FIRST. This dataset includes hierarchical and structured textual annotations, suitable for training text-controlled fashion generation models. Additionally, we have proposed two challenges on this dataset.
\item The preliminary quantitative and qualitative experiments indicate that FIRST can effectively enhance the generation quality of stable diffusion for fashion and improve the control of text over the generated images.
\end{itemize}
\begin{figure}
    \centering
    \includegraphics[width=0.90\linewidth]{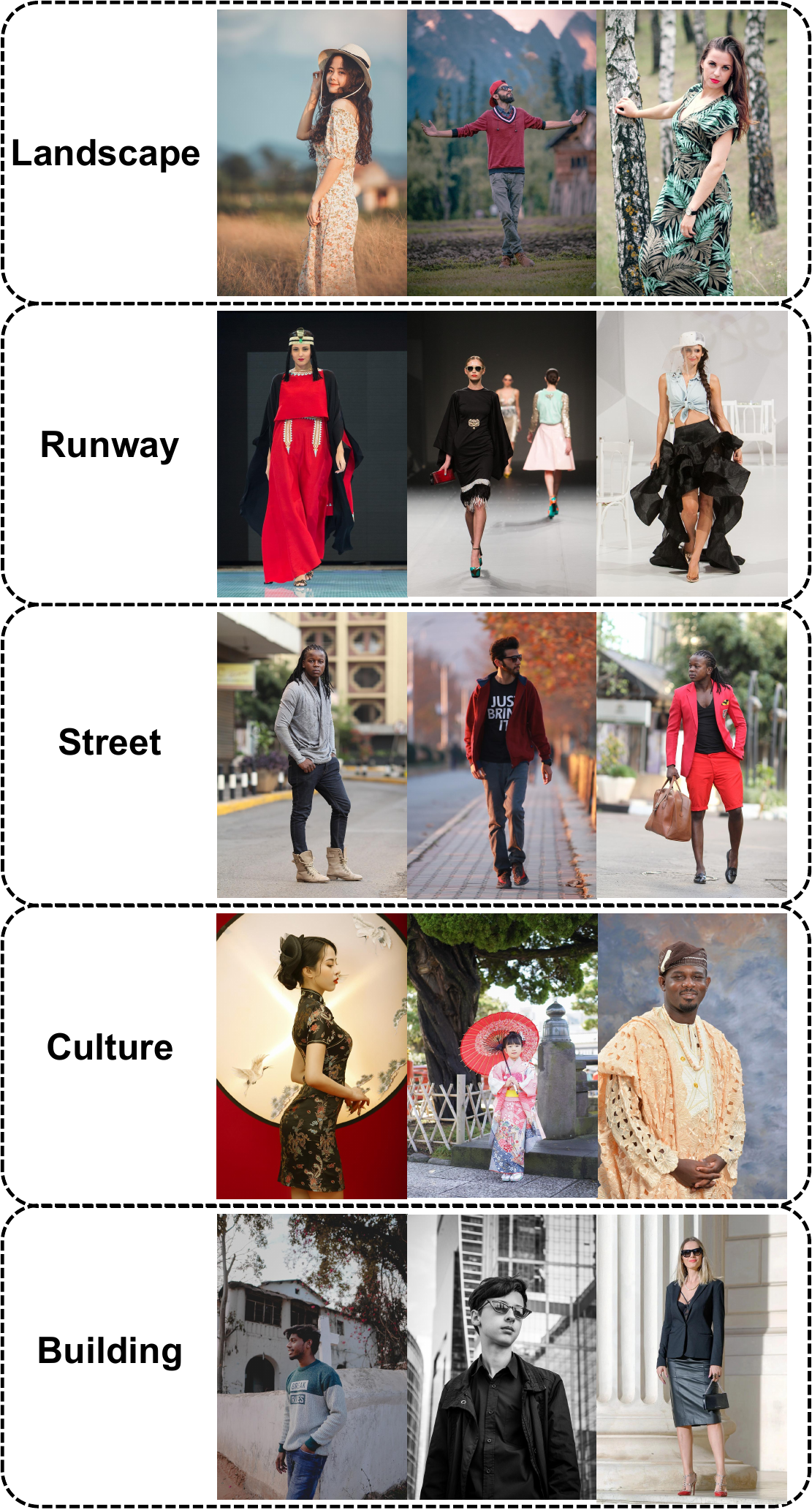}
    \caption{The examples of different photographic scenes.}
    \label{exa}
\end{figure}
\section{Related Work}
\subsection{Difusion Models}
Recently, diffusion models\cite{ho2020denoising,croitoru2023diffusion,nichol2021improved} have surfaced as an influential subset of generative models with a broad spectrum of applications across data modalities such as images, text, and audio\cite{leng2022binauralgrad,meng2021sdedit,su2022dual}. These models utilize a process that adds noise to an image over several iterations and then reverses it through denoising steps. They have been applied in various fields like local image editing\cite{lugmayr2022repaint,song2020score}, conditional generation\cite{choi2021ilvr,voynov2023sketch,zhang2023adding}, and image translation\cite{saharia2022palette,seo2023midms,tumanyan2023plug}. Recent innovations allow for task-specific control in diffusion models by pre-training them on personal devices\cite{shang2023post,li2023snapfusion,du2023exploring}, presenting a potential shift from training on large-scale GPU clusters. However, these methods may not yield the highest image quality for fashion images when multiple images are imposed as conditions, as they are not trained specifically on large-scale fashion image-text pairs.
\subsection{Text-to-image Synthesis}
While numerous studies have concentrated on crafting high-resolution images via diffusion models, a burgeoning segment of the research community is pivoting towards a more controlled generation process. Hertz and colleagues\cite{hertz2022prompt} explored a Prompt-to-Prompt technique for text-to-image creation, which utilizes text-driven activations within feature maps via cross-modal attention. The InstructPix2Pix\cite{brooks2023instructpix2pix} initiative integrates the vast pre-trained language capabilities of GPT-3 \cite{brown2020language} with the cutting-edge text-to-image LDM \cite{rombach2022high} to generate datasets geared towards text-influenced image manipulation. Despite these approaches' capacity to generate semantically aligned images, their reliance on extensive, open-domain datasets makes them less adept at discerning fashion-specific terminology.
\subsection{Fashion Synthesis}
Fashion synthesis and design is an emerging field within computer vision that deals with generating and manipulating fashion images, including clothing, accessories, and fashion figures. This research mainly branches into two directions: Virtual Try-On \cite{brown2020language,xu2021virtual,kim2019style,lewis2021tryongan} technologies that map new garment textures onto human figures using parsing and pose estimation, and garment-centric\cite{jiang2021deep,ding2023personalized,zhang2022armani,yu2019personalized} approaches that focus on creating new clothing items and styles. While VTON has been successful in aligning clothing attributes to human figures, garment-centric synthesis aims at generating diverse clothing styles, with current methods providing limited control over detailed attributes. Although text-to-image synthesis for fashion, which allows for editing garments based on textual descriptions, is in its nascent stages, it offers a new dimension of control but still lacks in achieving precision in texture and style details.
\subsection{Image Captioning}
Image captioning models are divided into two categories: supervised and unsupervised, depending on the availability of image-text alignment information during training. Supervised models\cite{xu2015show,cornia2020meshed,huang2019attention,chen2017sca} are trained with images and corresponding texts that align well, often using an encoder-decoder structure. They start by extracting visual features with vision backbones (\textit{e.g.}CNN\cite{he2016deep} or ViT\cite{dosovitskiy2020image}), which are then used by a language decoder (\textit{e.g.} LSTM\cite{hochreiter1997long} or Transformer\cite{vaswani2017attention}) to create coherent sentences. These models often incorporate attention mechanisms\cite{lu2017knowing, you2016image} to enhance vision-language alignment. Despite their effectiveness, the extensive data collection required for paired image-text data restricts their practical use. In contrast, unsupervised captioning\cite{feng2019unsupervised,laina2019towards,meng2022object} models use unrelated image and text data, leveraging visual concepts to form a makeshift alignment between the two. Our method distinguishes itself by requiring only text for training, bypassing the need for image processing during this phase. This approach not only cuts down on data-gathering costs but also proves to be more efficient than previous methods.
\begin{figure*}[]
\includegraphics[width=0.98\linewidth]{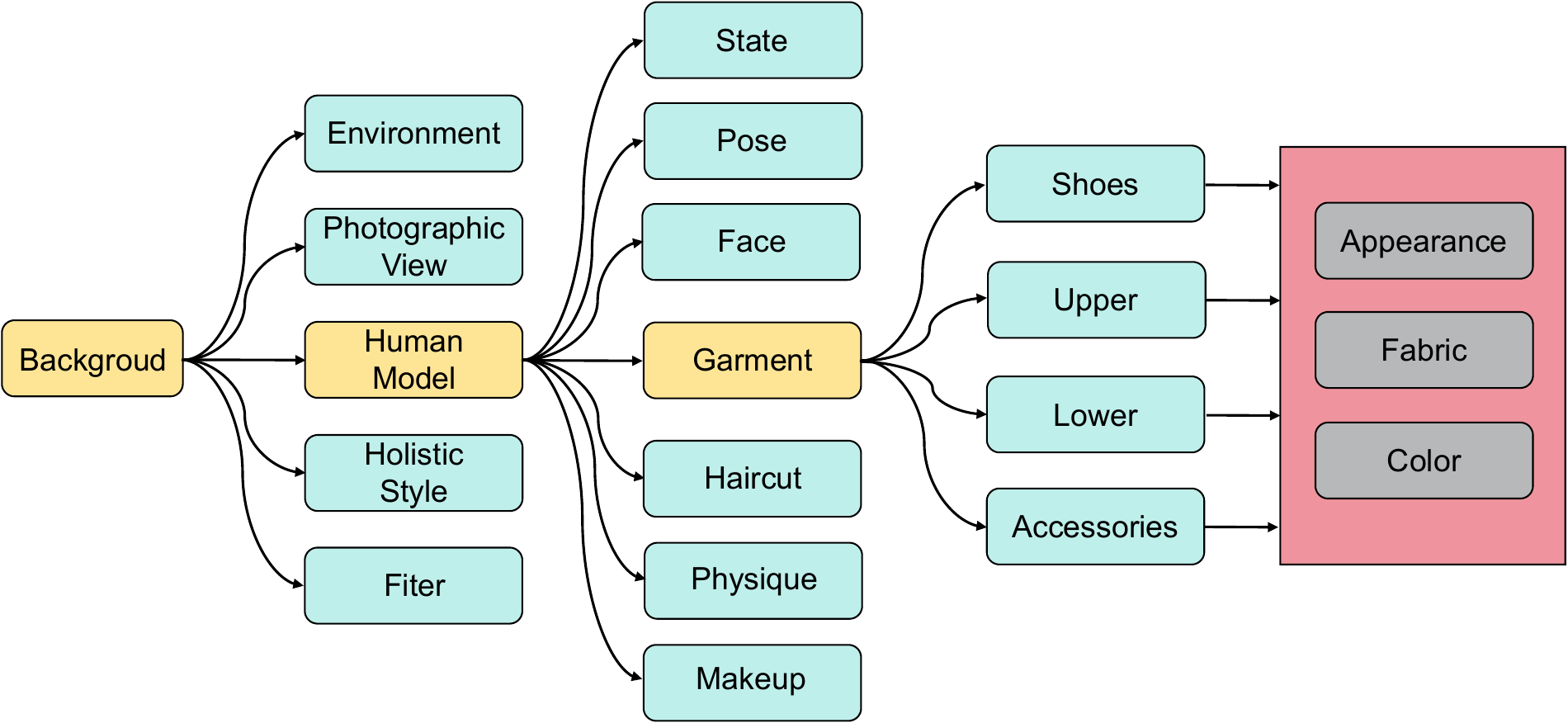}
\caption{
We describe fashion images according to the hierarchical structure shown in the diagram.
}
\label{hierar}
\end{figure*}

\section{FIRST Dataset}

In this section, we provide an overview of our proposed FIRST. We discuss it from four different aspects: data collection, data cleaning, data annotation, and data attributes, each of which takes much effort and enormous human resources. By walking through the above four facets, the details of the FIRST are present. At last, we compare FIRST with other fashion datasets and delve into their difference.
\subsection{Data Collection}
There are two main sources for our raw images. One source is from the Internet. We have designed and developed distributed web robots that are capable of crawling large numbers of fashion images from publicly accessible media sites, search engines, and fashion websites. Note that when crawling images, we verify the legitimacy of data crawling for those websites and the legitimacy of the images to ensure that the images are free to use for academic research. Another source is apparel manufacturers who have commercial partnerships with us. They provide the most high-quality and high-resolution fashion images. We also get their permission to utilize these images for research purposes. The total number of raw images is 1,124,371.
\begin{figure}
    \centering
    \includegraphics[width=0.90\linewidth]{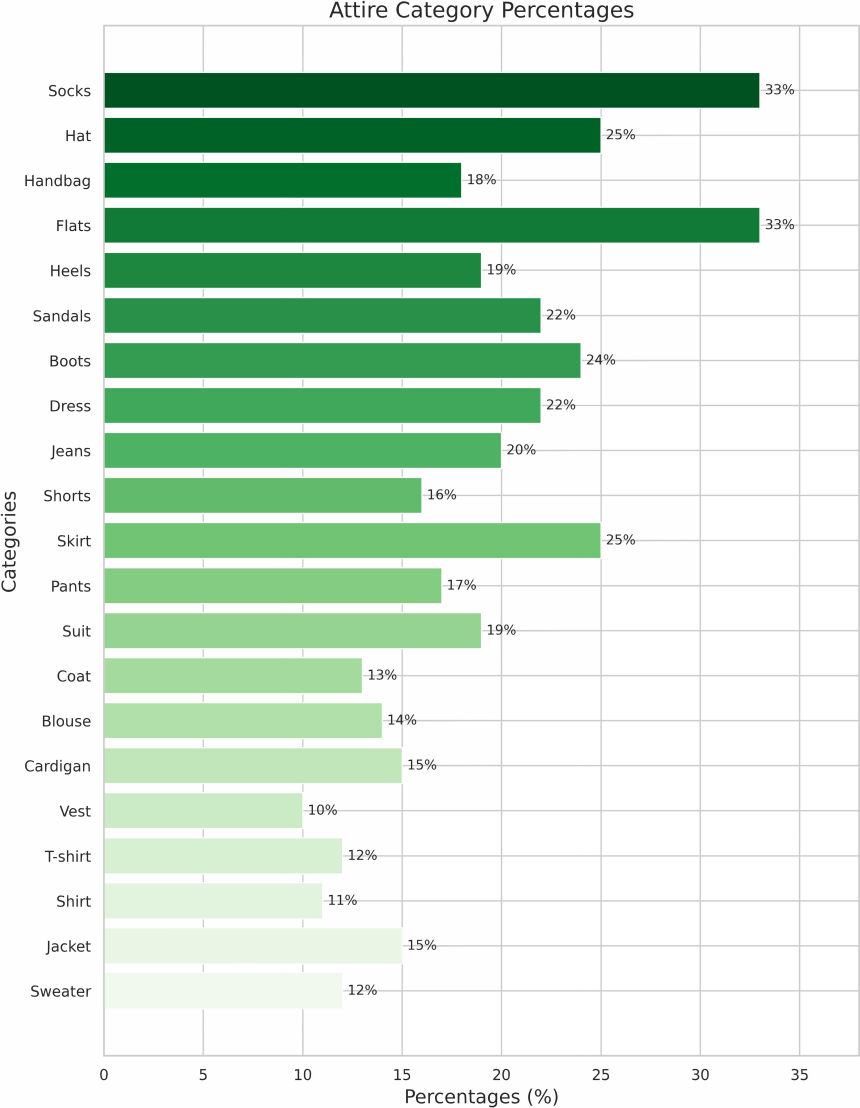}
    \caption{The distribution of different categories. Note that an image may include several categories, so the summation is not 1.}
    \label{category}
\end{figure}

\subsection{Data Cleaning}
A portion of the raw images are obtained from the Internet, and this portion contains a total of 482,339 images. Some of these images are extremely low-quality so they can not be used to train generative models. For example, a part of the images are very low-resolution, and some of them even contain visible watermarks. This part of the images affects the quality of generation and increases the difficulty of image labeling. To improve the overall quality of our constructed dataset, further cleaning of the raw images is required. The principles of data cleaning include (1) discarding images with a resolution lower than 512x512 (2) discarding images with watermarks (3) discarding images that are not related to fashion content (4) discarding images where fashion is too small. This labor-intensive work of data cleaning is done by humans. We employ 100 college students and dispatch the images to each of them. When the cleaning is completed according to the criteria, the remaining images are returned to us. After cleaning, the total of images is 1,003,451.

\subsection{Data Annotation}
\begin{figure*}[]
\includegraphics[width=0.98\linewidth]{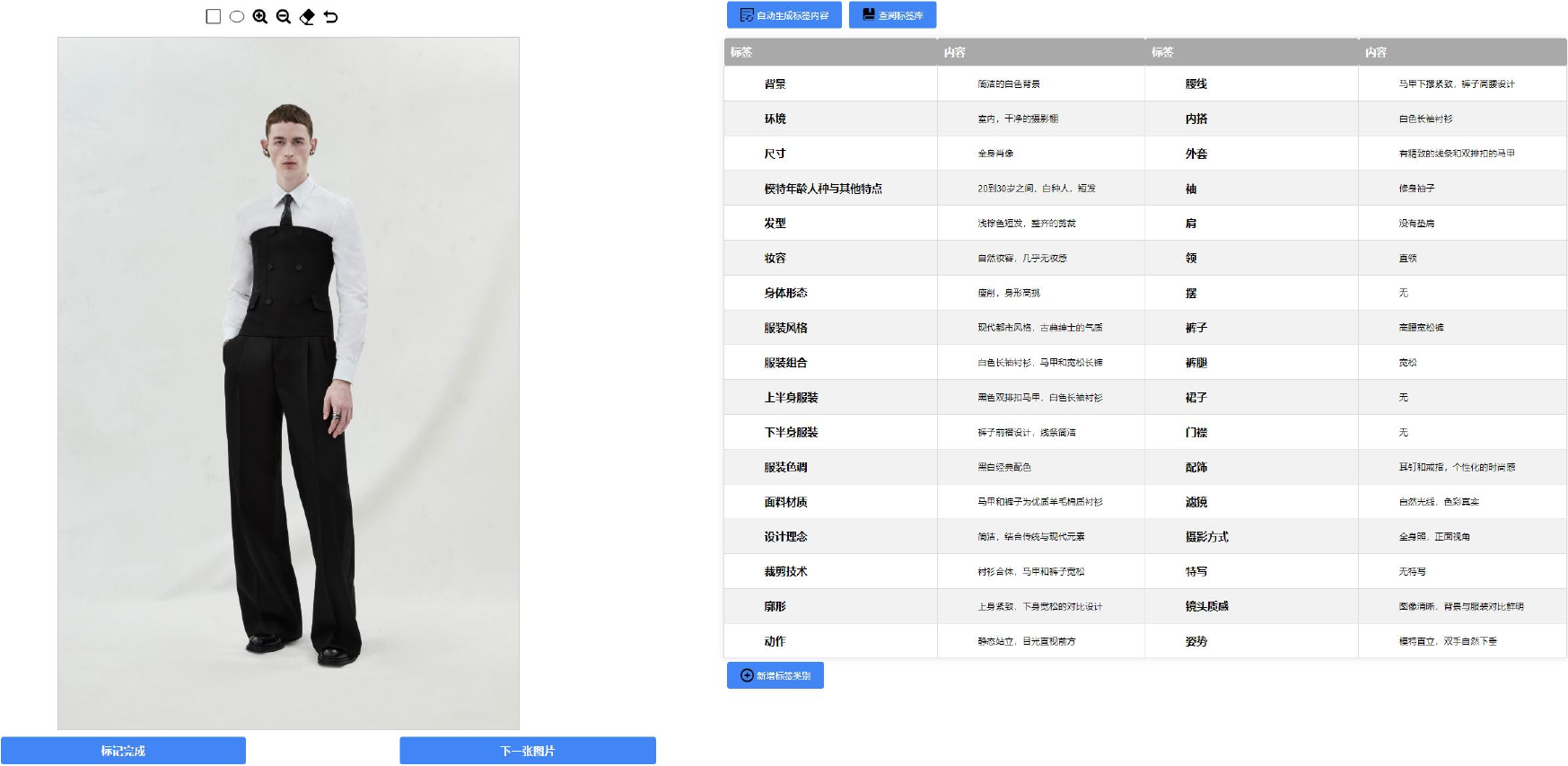}
\caption{
The web tool is comprised of two parts. The left part shows fashion images and the right part shows labels and their contents. The initial content is occupied with the texts from GPT-4V. Annotators can edit the initial content by their observations.
}
\label{webtool}
\end{figure*}
Captioning every image in our dataset is the most painstaking task, which consumes many human resources. It is impossible to elaborately describe a million fashion images in the same textual structure by human writing within a limited time, so we combine GPT-4V and human writing. Firstly, we define a particular prompt to guide the GPT-4V to generate a hierarchical textual description for each Image. Considering both the background and the models play a crucial role in fashion shows, together creating a comprehensive visual and emotional experience that helps convey the designer's creative concept and the essence of the fashion, it is necessary to add the text of the background and models to final annotations. Specifically, we induce GPT-4V to describe images from three levels: background, models, and garments.  For the background, GPT-4V captures the details and a holistic style of the environment where the image is taken. After that, GPT-4V starts to describe the appearance and physique of the model in the images. At the last level, GPT-4V focuses on the information of the garments in a coarse-to-fine manner. The coarse description includes color, categories, texture, material, and shape, while the fine description pays attention to small accessories and items. However, GPT-4V can not guarantee the correctness and correspondence of each text. we also need to revise the generated content. To do that, we employ 500 college students specializing in fashion design to check the annotation. Each student is assigned 2,007 images with paired texts. They read texts and align them with images. It takes 40 days to finish The whole process.
Figure \ref{webtool} shows the tool that students use to revise the annotation. The web tool uses Chinese as the default language. We translate the Chinese text into English text.

\subsection{Data Attributes}
Our dataset has four distinct attributes: Balanced Attire Categories, Multiple Photography Scenes, Hierarchical Annotation, and Collections, each of which is described below.

\textbf{Balanced Attire Categories.}
The richness of attire categories is important for fashion synthesis and design tasks. FIRST contains 21 different clothing categories such as sweaters, jackets, skirts, handbags, boots, and so on. The richness of the categories provides more learning templates for generative models, thus enhancing their creativity. The category distribution of these attire categories is shown in Figure \ref{category}. The figure shows that the majority of categories have a similar order of magnitude in total amount. Such a balanced distribution allows generative models to equally learn the concepts of each category and avoid the domination by minor categories during the training process. Based on the FIRST, we can also easily construct a long-tailed category distribution in the few-shot/long-tailed learning setting.

\begin{figure}[b]
    \centering
    \includegraphics[width=0.80\linewidth]{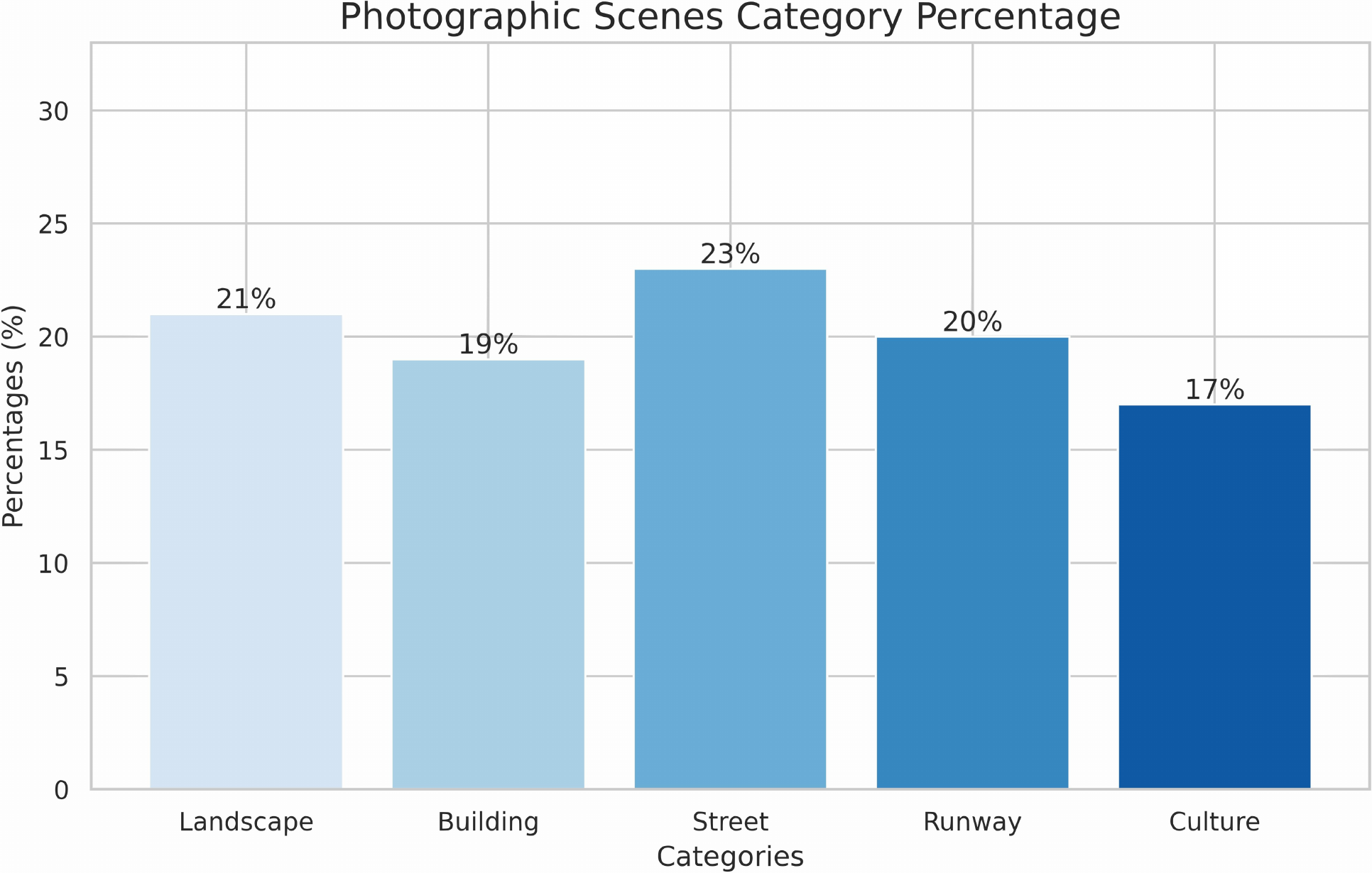}
    \caption{The distribution of different photographic scenes.}
    \label{scene}
\end{figure}

\textbf{Multiple Photographic Scenes.}
Generally, fashion images are shot in different scenes according to the core concept of design, and the images in our FIRST are also comprised of different photographic scenes. In the FIRST, the photographic scenes can be divided into 5 categories: landscape, building, street, runway, and culture, each of which contains a large number of fashion images with different backgrounds. Figure \ref{exa} shows examples of each scene. Figure \ref{scene} show the distribution of all types of scene. These categories cover most of the major fashion photographic backgrounds and satisfy the practical demand that designers desire diverse show backgrounds. For a versatile fashion design system, in addition to being able to generate models and garments, it should also be able to adaptively match the garments and models with appropriate photographic backgrounds to make the fashion display more reasonable. Our FIRST dataset provides the foundation to fulfill this.

\textbf{Hierarchical Annotation.}
Our hierarchy for annotation is shown in Figure \ref{hierar}. We transform the description of fashion images gradually from their global background to local fashion items. The philosophy of this hierarchy is to obtain a well-organized dataset. For backgrounds, we mainly focus our interest on the environment, holistic style, and filter. For the models, in addition to their appearance and physique, we also take into account their pose and state which are also an important part of the fashion exhibition. For garments, we follow a similar hierarchy. We first describe garments in a grand view which includes fashion styling, color scheme, textile, and philosophy of fashion design. Then, we continue to describe top clothes, pants, and shoes from a holistic view. We lastly extract the features of pockets, sleeves, or other small accessories. Using the above rules, we pair each image with a text and get a total labels of 1,003,451. Fig shows the hierarchy and 
examples of our text labels. To our best knowledge, such large-scale elaborate textual descriptions are the first work.

\textbf{Collections.}
To encourage the community to design algorithms or systems that are capable of creating a collection of fashions, we preserve the information about collections by filenames. We rename each image file with its publication date,  name of the designer, publisher, and unique collection ID. Fashion images in the FIRST can be grouped into 22,299 collections. Fig shows a few collections, and we can find that each collection follows the same visual pattern.
\subsection{Comparison}

\begin{table*}[ht!]
\centering
\caption{Dataset Features Comparison}
\label{hhh}
\scalebox{0.80}{
    \begin{tabular}{|l|c|c|c|c|c|}
    \hline
    Dataset & Total Image & Text Labels & Hierarchical Text Labels & Rich Photographic Scenes & Collection Classification \\ \hline
    Deepfashion\cite{liu2016deepfashion} & 146,680 & \xmark & \xmark & \cmark & \xmark \\ 
    Deepfashion-multimodal\cite{jiang2022text2human} & 44,096 & \cmark & \xmark & \xmark & \xmark \\
    SHHQ\cite{fu2022stylegan} & 231,176 & \xmark & \xmark & \cmark & \xmark \\
    VITON\cite{han2018viton} & 16,253 & \xmark & \xmark & \xmark & \xmark \\
    VITON-HD\cite{choi2021viton} & 13,679 & \xmark & \xmark & \cmark & \xmark \\ \hline
    FIRST & \textbf{1,003,451} & \cmark & \cmark & \cmark & \cmark \\ \hline
    \end{tabular}

}

\end{table*}

The landscape of fashion-related datasets is rich and varied, with numerous collections designed to serve a wide array of tasks within the domain of computer vision. To the best of our knowledge, our proposed FIRST is currently the largest dataset in scale compared to other datasets and has the richest textual annotations. 

In the dataset scale of existing open-source fashion datasets, the dataset volume of FIRST is 4.4 times bigger than that of SHHQ\cite{fu2022stylegan} (231,176) and is much larger than that of others (7 times to DeepFashion\cite{liu2016deepfashion}, 25times to DeepFashion-Multimodal\cite{jiang2022text2human} and 61 times to VITON\cite{han2018viton}).

In textual annotations, DeepFashion-MultiModal\cite{jiang2022text2human} is the most similar dataset to our FIRST among all fashion datasets, which is a large-scale high-quality human dataset with rich multi-modal annotations, containing 44,096 high-resolution human images. It also provides a textual description for each image. However, Deepfashion-Multimodal\cite{jiang2022text2human} only describes clothing length and cloth textures, while FIRST not only includes fashion styles, color, and philosophy of cloth but also depicts model and global background, providing more comprehensive fashion knowledge.

In terms of photographic scenes, images in FIRST are shot in various scenes, whose richness is larger than other fashion datasets. In contrast, DeepFashion-MultiModal\cite{jiang2022text2human}, VITON-HD\cite{choi2021viton} simply uses a pure white background. SHHQ\cite{fu2022stylegan} collects images shot in different scenes, but still less diverse than our proposed FIRST.

The most distinctive feature of our dataset is the meticulous categorization of garments according to their respective collections, a structural organization that is conspicuously absent in any other fashion datasets, which are often less orderly arranged. This deliberate and systematic classification not only reflects the inherent structure of fashion design but also sets the stage for the development of models capable of automatically generating cohesive collection-based fashion lines, thereby advancing the automation frontier in the fashion industry.

In summary, our FIRST proposes the largest amount of fashion images with the most diverse background, which also offers unique and hierarchical textual annotations among all fashion datasets. Table \ref{hhh} summarizes the difference of these datasets.

\section{Our Challenge}
The foremost challenge presented is the enhancement of diffusion models to accommodate inputs of extended textual length. Within our dataset, each text descriptor comprises nearly a thousand tokens to encapsulate detailed nuances. Notwithstanding, prevailing diffusion-based methodologies are constrained by an inherent limitation on token capacity due to their reliance on the CLIP architecture, which stipulates a maximum token count of 77. Such a limitation is markedly inadequate for the granularity required in the fashion domain, where details matter. Although the SDXL model attempts to circumvent this limitation by employing dual CLIP encoders, it still falls short of fulfilling the comprehensive needs of fashion industry applications, where extended descriptive capacity is essential. Thus, advancing these models to process longer text inputs remains an exigent need for bridging the gap between current capabilities and industry requisites.

The second challenge is the synthesis of cohesive fashion collections that are derived from a singular design ethos and inspiration.  In the realm of fashion design, a "collection" refers to an assembly of garments that, while varying in garment category, resonate with a shared aesthetic and conceptual narrative. This amalgamation of distinct fashion items, unified under the umbrella of a designer's unique vision, is profoundly abstract and often eludes even some fashion professionals.  Identifying and encapsulating the essence of a collection's inspiration poses a considerable hurdle, as it requires not only an understanding of the individual items but also an appreciation of the subtle interplay of themes and elements that bind them. By enabling fashion design systems to autonomously generate such collections, we could alleviate a significant portion of the intensive manual effort of fashion cloth designers involved in the design process. However, this task is fraught with complexity due to the intricate nature of translating abstract design inspirations into tangible fashion items that maintain a coherent visual and stylistic language across the collection.
\begin{figure}[b]
    \centering
    \includegraphics[width=0.99\linewidth]{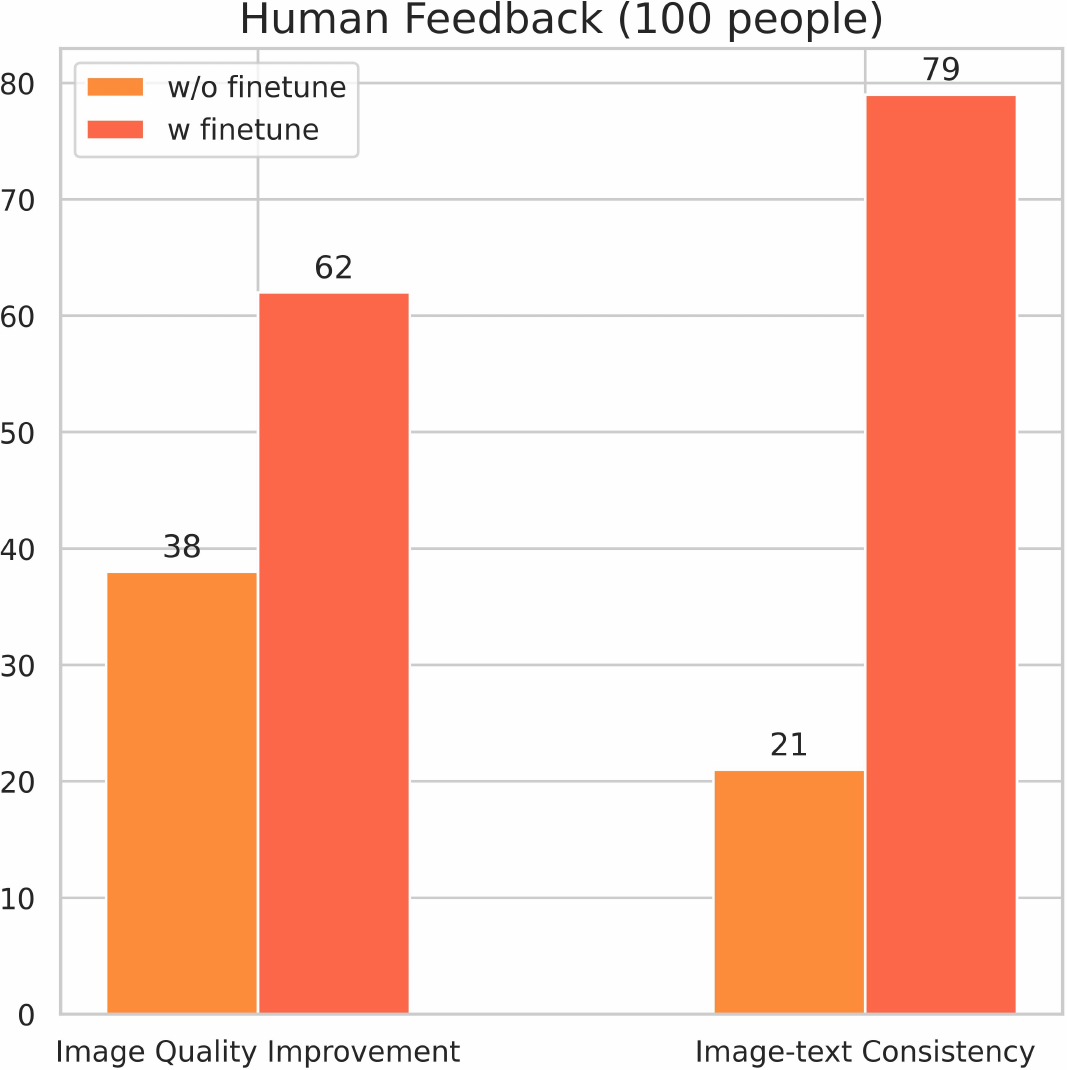}
    \caption{The results of human feedback. More volunteers think our generation quality is better than stable diffusion without fine-tuning on the FIRST.}
    \label{hf}
\end{figure}

\section{Experiment}
To validate the advancement and necessity of our dataset and illustrate the significance of the challenges we pose, we fine-tune stable diffusion using our data and analyze the generated results, highlighting the deficiencies therein. On the other hand, we distribute the images generated by stable diffusion before and after fine-tuning to volunteers who are responsible for assessing which generation has higher quality. Note that these volunteers all have a certain level of expertise in fashion design.
\subsection{Implementation}
We chose the stable diffusion model as our base due to its proven effectiveness in high-quality image generation. The model architecture is not modified; however, we update the hyperparameters to better suit our dataset characteristics. The dataset is split into training (80\%) and validation (20\%). We fine-tune the pre-trained stable diffusion model on our fashion dataset for 50 epochs using the AdamW optimizer with a learning rate of 1e-5, which is decreased by a factor of 0.1 every 15 epochs. A batch size of 256 is used due to GPU memory constraints. Our fine-tuning leverages a denoising objective tailored to our image domain, emphasizing texture and fine detail preservation. Our implementation was conducted using the PyTorch 2.0 deep learning framework.
\begin{figure*}[]
\includegraphics[width=0.98\linewidth]{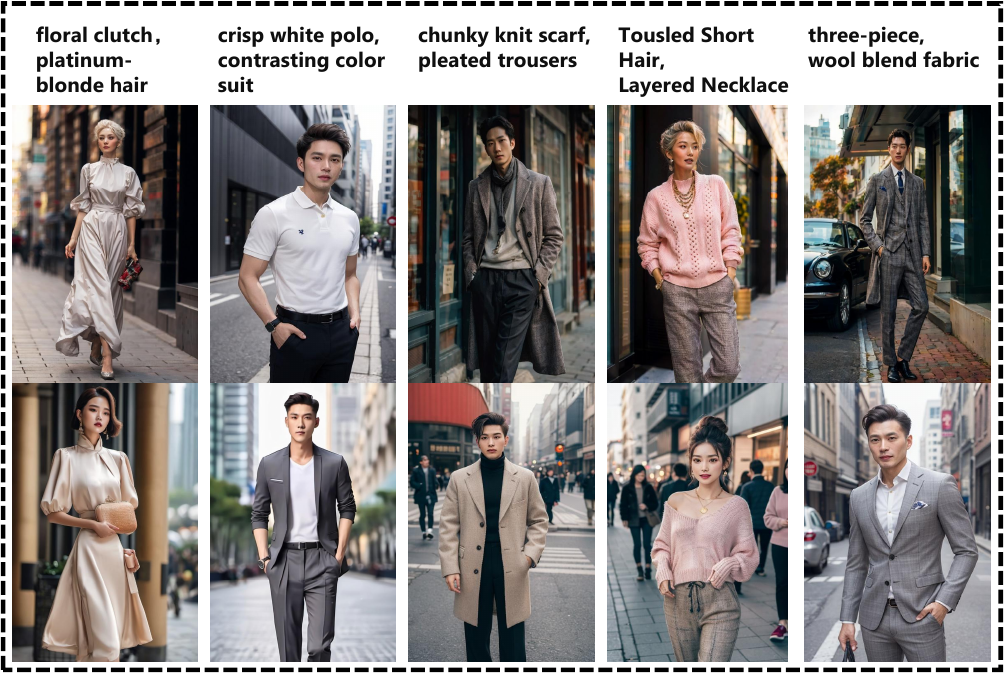}
\caption{
Comparison between stable diffusion models before and after fine-tuning. The first row indicates parts of the input prompt. The second row represents the images after being fine-tuned on the FIRST. The third shows the images from the original stable diffusion model. 
}
\label{quan}
\end{figure*}

\subsection{Quantitative \& Qualitative Result}
In our study, we adopt two metrics: FID (Fréchet Inception Distance) and CLIP-S (CLIP Score). FID serves as an indicator of the image generation quality, while CLIP-S reflects the control ability of the input language over the generated images.
Table \ref{tab1} shows the FID and CLIP-S(CLIP score) of stable diffusion before and after fine-tuning over the FIRST. we can conclude that 
 after fine-tuning the stable diffusion model on our dataset, there is a pronounced improvement in performance on the validation set. At a resolution of 256x192, the FID decreases by 12.36 while the CLIP-S increases by 3.81; at a resolution of 512x384, the FID decreases by 12.41, and the CLIP-S increases by 3.48. This optimization suggests that a customized dataset is essential for fashion synthesis and design, as it leads to improvements in both generation quality and language control. We also visualize the sampling results of the two models at a resolution of 512x384 in Figure \ref{quan}. We can find that the original stable diffusion does not capture details from the input text well.
\begin{table}[]
\centering
\caption{Result Comparison}
\label{tab1}
\scalebox{1.1}{
\begin{tabular}{cccc}
\hline
\textbf{Model} & \textbf{Resolution} & \textbf{FID}   & \textbf{CLIP-S} \\ \hline \hline
w/o finetune   & 256x192             & 20.19          & 18.74           \\ 
w finetune     & 256x192             & \textbf{7.83}  & \textbf{22.55}  \\ \hline
w/o finetune   & 512x384             & 22.73          & 19.39           \\ 
w finetune     & 512x384             & \textbf{10.32} & \textbf{22.87}  \\ \hline
\end{tabular}
}
\end{table}
\subsection{Human Feedback}
We also distribute the results generated by different models before and after fine-tuning to a randomly selected 100 volunteers, with each person receiving a distinct pair of images. Both images in each pair are generated under the control of the same text prompt. The volunteers are unaware of the correspondence between images and models, and they are asked to select the image that better matches the text description and the one with higher generation quality. We conduct a statistical count of the selections. The results are shown in the Figure \ref{hf}. It can be observed that 79 people believe that the images generated by the fine-tuned model match better with the input text, while 62 people find the quality of generation improved.

\section{Conclusion}
This paper presents a new large-scale fashion dataset FIRST containing one million diverse fashion images and rich hierarchical text annotations. Preliminary experiments show that these image-text pairs can be used to improve the quality of diffusion-based generative models for generating garments and enhance text-to-image capability, which can facilitate the development of advanced garment synthesis and design systems. Based on the FIRST, we also present two challenges: the long text prompt problem and garment collection generation. The two challenges can inspire the community to develop more powerful garment designers.
In the future, we will improve our existing work in two aspects. On the one hand, we will continue to increase the scale of the dataset and further improve the quality of image annotation. Also, we plan to enrich the diversity of the data. On the other hand, we will design a simple and effective baseline model for the proposed challenges as a reference for the community.

{\small
\bibliographystyle{ieee_fullname}
\bibliography{egbib}
}

\end{document}